\documentclass[11pt,a4paper]{article}
\usepackage[hyperref]{acl2020}
\usepackage{times}
\usepackage{booktabs}
\usepackage{latexsym}
\usepackage{xspace}
\usepackage{tabularx}
\usepackage{graphicx}
\usepackage{multirow}
\usepackage{amsfonts}
\usepackage{multicol}
\usepackage{amsmath}
\usepackage{multirow}

\aclfinalcopy 

\usepackage{microtype}

\newcommand{\ie}{i.e.,\xspace}
\newcommand{\eg}{e.g.,\xspace}
\newcommand{\eat}[1]{}
\newcommand{\paratitle}[1]{\noindent\textbf{#1}}

\newcommand{\baby}{\textsc{PPVAE}\xspace}
\newcommand{\glob}{\textsc{PretrainVAE}\xspace}
\newcommand{\cond}{\textsc{PluginVAE}\xspace}
\newcommand{\globen}{pre-trained global encoder\xspace}
\newcommand{\globde}{pre-trained global decoder\xspace}
\newcommand{\ours}{\xspace\textit{(ours)}}

\def\aclpaperid{806} 

\title{Pre-train and Plug-in: Flexible Conditional Text Generation with Variational Auto-Encoders}
  
  \author{Yu Duan$^1$\thanks{\ \ \ The first three authors contribute equally to this paper.} , Canwen Xu$^2$$^*$, Jiaxin Pei$^3$$^*$, Jialong Han$^4$\thanks{\ \ \ Work done when Jialong Han was with Tencent AI Lab.}, Chenliang Li$^2$\thanks{\ \ \ Chenliang Li is the corresponding author.}\\
 $^1$ Alibaba Group, China $^2$ Wuhan University, China \\
 $^3$ University of Michigan, United States $^4$ Amazon, United States \\
 $^1$ {\tt derrick.dy@alibaba-inc.com}, $^2$ {\tt \{xucanwen,cllee\}@whu.edu.cn} \\ $^3$ {\tt pedropei@umich.edu}, $^4$ {\tt jialonghan@gmail.com}
}
 
\begin{document}
\maketitle
\begin{abstract}
Conditional Text Generation has drawn much attention as a topic of Natural Language Generation~(NLG) which provides the possibility for humans to control the properties of generated contents. Current conditional generation models cannot handle emerging conditions due to their joint end-to-end learning fashion. When a new condition added, these techniques require full retraining. In this paper, we present a new framework named \textbf{P}re-train and \textbf{P}lug-in  \textbf{V}ariational \textbf{A}uto-\textbf{E}ncoder~(\baby) towards flexible conditional text generation. \baby decouples the text generation module from the condition representation module to allow ``one-to-many'' conditional generation. When a fresh condition emerges, only a lightweight network needs to be trained and works as a plug-in for \baby, which is efficient and desirable for real-world applications. Extensive experiments demonstrate the superiority of \baby against the existing alternatives with better conditionality and diversity but less training effort.\footnote{The code is available at \url{https://github.com/WHUIR/PPVAE}.}
\end{abstract}

\section{Introduction}
Currently, neural generation techniques have powered many inspiring applications, \eg poem generation~\cite{Yang:18}, neural machine translation (NMT)~\cite{Bahdanau:14} and chatbot~\cite{Zhao:17}. Conditional (also known as controllable) text generation is an important task of text generation, aiming to generate realistic text that carries a specific attribute (\eg positive or negative sentiment). A common solution is to encode the condition into a vector representation and then integrate it with the text generation process~\cite{Kingma:14a,Hu:17,Mirza:14}. 
These existing neural models have achieved encouraging results. However, when a new condition is added (\eg a new topic for categorical generation), they require a full retraining or fine-tuning. This process is both time-consuming and computationally inefficient~\cite{icml:adapter}. Both fine-tuning and retraining are not desirable in real-world applications since the delivery (\eg transmitting updated weights through the Internet) and client-side re-deployment (\eg distribute updated weights to users) of large-scale weights are often difficult.

Inspired by the recent success of Variational Auto-Encoder (VAE)~\cite{Kingma:13} based post-hoc conditional image generation strategy~\cite{Engel:17}, we provide a new perspective for flexible conditional text generation. We propose \textbf{P}re-train and \textbf{P}lug-in  \textbf{V}ariational \textbf{A}uto-\textbf{E}ncoder (\baby), which decouples the text generation module from the condition representation module. \baby is a hierarchical framework composed of two VAEs: \textbf{(1) \glob}, which derives the global latent space of text with its encoder (\globen) and learns to generate text based on an easily-accessible large unlabeled dataset with its decoder (\globde); \textbf{(2) \cond}, which is a lightweight neural network that learns to transform vectors from the conditional latent space to the global latent space, and vice versa. This mapping function can be easily learned with only a few conditional training samples. In this sense, once we transform a latent variable (also known as latent code) randomly sampled from the conditional space distribution to the global space, the \globde is directly adopted for generation. In other words, whenever a new condition emerges, we only need to train a \cond and directly plug it into the framework.

Different from the existing end-to-end neural models~\cite{Mirza:14,Sohn:15,Kingma:14a}, \baby focuses on the learning of pure transformation between the continuous latent spaces, instead of the tricky discrete text generation. Once trained, \glob is fixed for text representation and generation under all conditions.
Our proposed framework decouples the conditional space learning from the text generation, endowing \baby with more flexibility when handling emerging conditions. Also, training only a small conditional network for latent space transformation is much more efficient than co-training with the text generation. Additionally, we can easily increase the capability of generation using a larger corpus or deeper neural networks for text encoding and decoding. Our main contributions can be summarized as follows:
(1) We propose a novel framework, \baby, for conditional text generation, which allows a separate training for a new condition without retraining the whole network. (2) We conduct extensive experiments and analysis to verify the effectiveness of our proposed \baby. Our framework achieves state-of-the-art performance on conditionality in both automatic and human evaluations.

\section{Related work}
Boosted by the recent success of deep learning technology, Natural Language Generation~(NLG) has recently become popular in the NLP community. Many great works have attempted to solve various subtasks like dialogue generation~\cite{Li:16}, poetry generation~\cite{Yi:18} and story generation~\cite{Fan:18} and new techniques keep emerging~\cite{Bowman:16,Yu:17,chunshu}.
However, due to the black-box nature of neural networks, the recent proposed generic models suffer the problem of lacking interpretability and controllability. 

To handle this problem and support generating plausible text with a specified condition, conditional text generation~\cite{Kikuchi:16,Ficler:17,Hu:17} has recently attracted extensive attention. Current research in this direction mainly falls into two fashions: the supervised methods and semi-supervised methods. For supervised methods, \citet{Mirza:14,Sohn:15} first converted the condition information to one-hot vectors, then integrated them into a generator and a discriminator. To enhance the correlation between structured conditional code and generated samples, \citet{Chen:16} adopted an extra adversarial classifier to infer the structured code from generated samples. \citet{Wang:18} used multiple generators for multiple conditions and a multi-class classifier to provide training signals for the learning of generators.

However, given only a limited number of conditional samples, semi-supervised methods are compulsory. To utilize the implicit conditional distribution behind the unlabeled text, \citet{Kingma:14a} introduced a classifier into the VAE architecture. \citet{Hu:17} further involved two additional independent regularization terms in enhancing the disentanglement between structured code and unstructured code. Very recently, \citet{ctrl} used human-defined ``control code'' to pre-trained Language Model in an unsupervised manner.

Our work falls in the category of semi-supervised learning yet differs from the existing works in the following ways: (1) Our model decouples the text generation module from the condition representation module which two are tightly fused as a single one in previous studies, enabling possible exploitation for pre-trained Language Models~(\eg GPT-2~\cite{gpt2}). (2) Our model allows single-condition generation, which could inspire new applications like polite speech generator~\cite{niu2018polite} and data augmentation~\cite{guoday1}. (3) Our model can handle emerging conditions while achieving state-of-the-art performance with fewer parameters and less training time.

\section{Preliminaries}

\paratitle{Variational Auto-Encoder (VAE).} VAE~\cite{Kingma:14b} is widely used in continuous generation (\eg image generation). \citet{Bowman:16} introduced VAE to NLG to solve the ``one-to-many'' generation problem (\ie generating multiple feasible samples for the same input). Given a latent variable $z$ randomly sampled from a prior distribution, VAE comprises an encoder $enc(x) = q_\phi(z|x)$ and a decoder $dec(z) = p_\theta(x|z)$. The encoder aims to encode input data $x$ into latent space $Z \in \mathbb{R}^d$. The decoder is used to reconstruct the original input $x$, given the corresponding $z$. Thus, the loss function of VAE is formulated as:
\begin{equation}
\begin{aligned}
	\mathcal{L}_{\mathit{VAE}}(x) = & - \mathbb{E}_{q_\phi(z|x)}[\log p_\theta(x|z)] \\ & + \operatorname{KL}(q_\phi(z|x)\|p(z))
\end{aligned}
\end{equation}
where $\operatorname{KL}(\cdot||\cdot)$ is the Kullback-Leibler (KL) divergence, $p(z) = \mathcal{N}(0, 1)$ is the prior distribution. The first term ensures that VAE can distill compact variable $z$ in latent space for reconstruction. The second term pushes posterior distribution to be close to the prior distribution, securing the mutual information between original data and the latent space~\cite{Dupont:18}.

\paratitle{Conditional Text Generation with VAE.} Conditional text generation has drawn much attention recently. By controlling the properties of generated contents, we can apply the generative models to many real-world scenarios. We follow the problem setting in \cite{Hu:17}. Given a set of $k$ conditions $C=\{c_1, c_2, ..., c_k \}$, an unlabeled corpus $X$, and conditional text samples $Y=Y_1 \cup Y_2 \cup ... \cup Y_k$ where each $Y_i$ is a set of text samples that carries the condition $c_i$. The goal of a VAE model is to learn a decoder $p_\theta(\hat{y}|z, c_i)$ that takes the latent variable $z$ and the condition $c_i$ to calculate the distribution over the text samples $Y_i$. Thus, when the condition $c_i$ and a randomly sampled latent variable $z \sim p(z)$ specified, the model could generate realistic text samples matching the given condition.

\section{Pre-train and Plug-in Variational Auto-Encoder}

\begin{figure*}
\centering
\includegraphics[width=\textwidth]{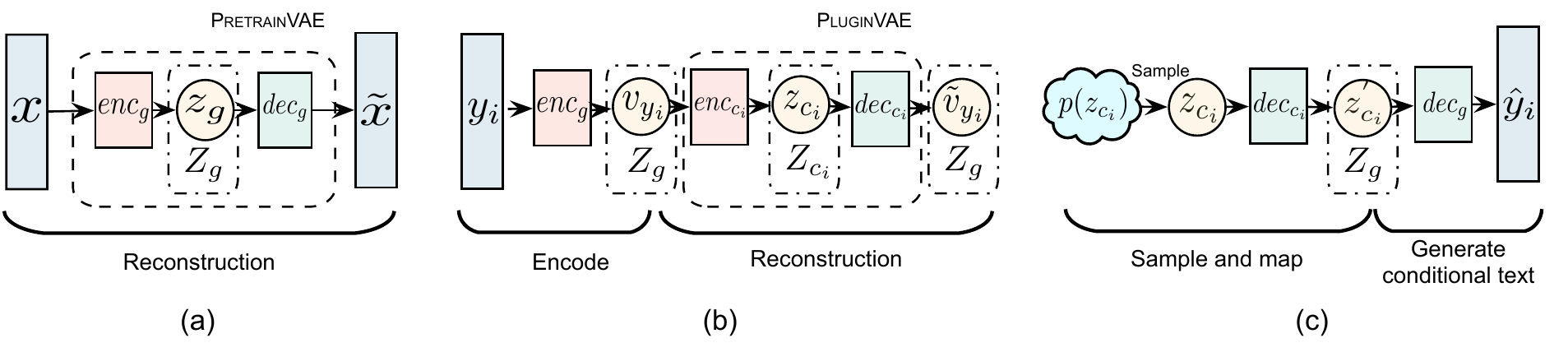}
\caption{The whole workflow of our proposed framework.}

\label{img-framework}
\end{figure*}

As a basis for semi-supervised learning, a large unlabeled corpus should include diverse text which covers a vast spectrum of conditions.
Thus, text under each condition forms a conditional latent space, which could be mapped from a larger global latent space.
Based on this, we propose a \glob and a \cond to derive the global and conditional latent space, respectively.
\subsection{Framework}

 \glob is composed of a pre-trained global encoder for text representation and a pre-trained global decoder for text generation.

\paratitle{\glob.} The encoder and decoder of \glob are used to encode and generate text, respectively.
As discussed above, \glob is trained on a large amount of unlabeled text to derive the global latent space $Z_g$ for the latent variable $z_g$, where $Z_g \in \mathbb{R}^{d_g}$ and $d_g$ is the space dimension. 
Previous studies usually use a common VAE for text representation and generation. However, as pointed out in \cite{Bowman:16}, VAE suffers the notorious ``posterior collapse'' problem. To address this, we utilize Wasserstein Autoencoder (WAE)~\cite{Tolstikhin:17} for \glob.
Different from the original VAE, WAE encourages aggregated posterior distribution to be close to the prior, which is effective in alleviating the reconstruction problem of VAE~\cite{Tolstikhin:17}. Specifically, we adopt WAE-GAN, a variant of WAE, which incorporates the merits of adversarial learning. During training, the encoder $enc_g(x) = q_g(z_g|x)$ encodes the text to the latent space and the decoder $dec_g(z_g) = p_g(x|z_g)$ reconstruct the text with the latent variable $z_g$. Thus, the loss function of \glob is formulated as:

\begin{equation}
\label{globalloss}
\begin{aligned}
\mathcal{L}_{\glob}(x) = & -\mathbb{E}_{q_g(z_g|x)}[\log p_g(x|z_g)] \\
& +\lambda D(Q(z_g),p(z_g))
\end{aligned}
\end{equation}
where $Q(z_g)= \int q_g(z_g|x)p(x)~dx $ is the aggregated posterior distribution; $p(z_g)$ is the prior normal distribution; $D$ is the adversarial discriminator; $\lambda$ is the coefficient hyper-parameter ($\lambda >0$).

\paratitle{\cond.} For each condition, we use a condition-specific \cond to derive the conditional space.
That is, \cond is proposed to learn the transformation between the conditional and global latent space for each condition. Specifically, for each condition $c_i$, we use a limited number of conditional samples $y_i$ and utilize the global encoder $enc_g$ to encode them into $v_{y_i}$. Note that normally, the encoded text samples under a single condition are not likely to densely clustered in the global text space $Z_g$, since the learning process of $Z_g$ is condition-independent and the unlabeled corpus contains diverse text samples. \cond for condition $c_i$ consists of an encoder $enc_{c_i}(v_{y_i})=q_{c_i}(z_{c_i}|v_{y_i})$ and a decoder $dec_{c_i}(z_{c_i})=p_{c_i}(v_{y_i}|z_{c_i})$. The learned condition-dependent latent space is $Z_{c_i} \in \mathbb{R}^{d_c}$, where $d_c$ is the space dimension. Thus, \cond is capable of mapping the samples in the global latent space to and from a denser conditional latent space (\ie $d_c < d_g$). During training, the loss function of \cond for a single condition is defined as:
\begin{equation}
\label{equ-condVAE}    
\begin{aligned}
\mathcal{L}_{\mathit{single}}(v_{y_i}) &= -\mathbb{E}_{q(z_{c_i}|v_{y_i})}[\log p_{c_i}(v_{y_i}|z_{c_i})]\\
&+\mid (\operatorname{KL}(q_{c_i}(z_{c_i}|v_{y_i}) \|p(z_{c_i})) - \beta \mid 
\end{aligned}
\end{equation}
where $p(z_{c_i})$ is the prior normal distribution of the conditional latent space; $z_{c_i}$ is the latent variable; $v_{y_i}=enc_g(y_i)$ is encoded text samples from $Y_i$. To enhance the diversity of generated text, we introduce an extra constant term $\beta$ to control the amount of encoded information in VAE~\cite{Dupont:18,ChenLGD:18,Kim:18}. By setting $\beta$ to an appropriate value, \cond could extract compact conditional information without sacrificing the fluency or accuracy. 

Although we can already generate conditional text under a single condition by Equation \ref{equ-condVAE}, it is possible to even further improve the conditionality by introducing negative samples. We construct the negative samples $y^{'}_i$ from $Y^{'}_i$ and encode them:
\begin{equation}
\begin{split}
Y^{'}_i = Y - Y_i \\
v^{'}_{y_i} = enc_g(y^{'}_i)
\end{split}
\end{equation}

Thus, the loss function of \cond with negative samples is defined as:
\begin{equation}
\label{multi-cond}
\begin{aligned}
&\mathcal{L}_{\cond}(v_{y_i}, v^{'}_{y_i}) \\
&= \mathcal{L}_{\mathit{single}}(v_{y_i})-\gamma~\mathcal{L}_{\mathit{single}}(v^{'}_{y_i})  
\end{aligned}
\end{equation}
where $v_{y_i}$ is a batch of encoded samples under condition $c_i$, and $v^{'}_{y_{i}}$ is a batch of encoded negative samples;
$\gamma$ is a hyper-parameter balancing the positive and negative samples.
For different tasks, the best setting for $\gamma$ may vary. Intuitively, the larger the difference between the conditions is, the smaller $\gamma$ should be.

\subsection{Workflow}
In this section, we provide the details of training and generation procedures.
As illustrated in Figure \ref{img-framework}, the workflow is composed of three steps.

\paratitle{Pre-train once, infer everywhere.} First, as shown in Figure \ref{img-framework}(a), using the unlabeled corpus $X$, we pre-train \glob to learn the global latent space $Z_g$ by reconstruction with Equation \ref{globalloss}. Once pre-trained, the weights of both $enc_g$ and $dec_g$ are fixed. As an unsupervised VAE model, \glob is capable of generating diverse but unconditional text. 

\paratitle{Train it when you need it.} Previous methods~\cite{Kingma:14a,Hu:17} learn the joint conditional space by jointly considering all conditions. However, once the model is trained, it is not possible to add a new condition without a full retraining. Different from those approaches, \baby is totally flexible that allows adding new conditions. Shown in Figure \ref{img-framework}(b), once a condition is added, we only need to train a \cond specifically for this condition with Equation \ref{equ-condVAE} (or Equation \ref{multi-cond}, if provided with samples of other conditions). Since \cond is text-irrelevant and only learns to map between two latent spaces, the training number of parameters is only $0.34\%$ (see Section \ref{traincosts}) of fine-tuning \glob or retraining other models.  Additionally, although we need to train $k$ \cond for $k$ conditions, the total number of trained parameters is still much smaller than existing methods (unless $k> 1/0.34\%\approx 294$, which is impossible in actual applications). Plus, we can parallel the conditional training to speed up the process easily.

\paratitle{Plug it in and generate.} Shown in Figure \ref{img-framework}(c), once \cond for the condition $c_i$ is trained, we can plug it into the \baby framework and generate text together with \glob.

First, we randomly sample a latent variable $z_{c_i}$ from the prior distribution $p(z_{c_i})=\mathcal{N}(0,1)$. Then we use \cond's decoder $dec_{c_i}$ to map $z_{c_i}$ to the global latent space $Z_g$ and obtain $z^{'}_{c_i}$:
\begin{equation}
 z^{'}_{c_i} = dec_{c_i}(z_{c_i}).   
\end{equation}
Since $z^{'}_{c_i} \in Z_g$, we can directly use the global decoder $dec_g$ to generate text:
\begin{equation}
  \hat{y}_i=dec_{g}(z^{'}_{c_i})
\end{equation}
where $\hat{y}_i$ is the generated text under condition $c_i$. 

\section{Experimental Settings}
\subsection{Datasets}

Following the setting of \cite{Hu:17}, we mainly focus on short text generation (no longer than 15 tokens), which is easier for both automatic and human evaluations.
We use Yelp~\cite{Shen:17} and News Titles~\cite{Fu:2018} for experiments. Yelp is a collection of restaurant reviews. We use the pre-processed version used in~\cite{Shen:17}, where two polarity sentiment labels are provided. For News Titles, we choose the titles belong to \textit{Business}, \textit{Entertainment} and \textit{Health} categories for our experiments.

Both Yelp and News Titles are datasets with relatively short text. We filter out text longer than 15 words, then choose the top 8,900 and 10,000 words as the vocabulary for Yelp and News Titles, respectively. The statistics of the two datasets are listed in Table~\ref{dataset-info}. We discard the labels in the original training and validation splits. We use the original training split as the unlabeled corpus; the validation split to select the best unsupervised models, and the test split as the labeled conditional text. 

Based on the Yelp dataset, we define two tasks: (1)~\textbf{Sentiment.} This task aims at generating text samples, either positive or negative. The ratio of positive/negative text in Yelp is roughly $0.6:0.4$. We randomly sample 200 positive and 200 negative text for supervised training. (2)~\textbf{Length.} This task aims at generating text samples with a specific length. We define ($len \leq 3$) as short text, ($len \geq 12$) as long text and ($3<len<12$) as medium text. We respectively sample 200 text for short, medium, and long text for supervised training.

Based on the News Titles dataset, we define the categorical text generation task called
\textbf{Topic.} This task aims at generating text samples on a certain topic. The ratio of business/health/entertainment in News Title is $0.38:0.15:0.47$, which is more imbalanced than Yelp. We randomly sample 200 text for each category for supervised learning.

\begin{table}[t!]
\small
\begin{center}
\begin{tabular}{ccccc}
\toprule
Dataset & \#Train & \#Dev & \#Test & Avg-len \\ 
\midrule

Yelp & 444,101 & 63,483 & 126,670 &  8.93 \\
News Titles & 249,043 & 23,949 & 20,000 & 9.85 \\

\bottomrule
\end{tabular}
\caption{\label{dataset-info} The statistics of Yelp and News Titles.}
\end{center}
\end{table}

\begin{table*}[t!]
\small
\begin{center}
\begin{tabularx}{\textwidth}{c|c|l|cc|cc}
\toprule
\multirow{2}{*}{Task} & \multirow{2}{*}{Conditions} & \multirow{2}{*}{Method} & Accuracy& Log-Variance & Distinct-1 & Distinct-2 \\ 
& & & ($\uparrow$ better) & ($\downarrow$ better) & ($\uparrow$ better) & ($\uparrow$ better)\\
\midrule
\multirow{4}{*}{Sentiment} & \multirow{4}{*}{\{Positive, Negative\}} & S-VAE & 0.7194 & -5.38 & 0.0198 & 0.2520  \\
&& CTRL-GEN & 0.6998 & -2.78 & 0.0026 & 0.0164 \\
&& \baby-single\ours & \underline{0.7832} & \underline{-11.12} & \underline{0.0350} & \underline{0.2568}\\
&& \baby\ours & \textbf{0.8484} & \textbf{-11.90} & \textbf{0.0356} & \textbf{0.2627} \\
\midrule
\multirow{4}{*}{Length} & \multirow{4}{*}{\{Short, Medium, Long\}} & S-VAE & 0.8598  & -4.82 & 0.0187 & 0.1795 \\
&& CTRL-GEN & 0.3957 & -1.96 & 0.0021 & 0.0146 \\
&& \baby-single\ours & \underline{0.9640} & \underline{-6.96} & \textbf{0.0375} & \textbf{0.2549} \\
&& \baby\ours & \textbf{0.9722} & \textbf{-7.64} & \underline{0.0372} & \underline{0.2538} \\
\midrule
\multirow{4}{*}{Topic} & \multirow{4}{*}{\{Business, Health, Entmt.\}}& S-VAE & 0.6930 & -2.32 & 0.0360 & 0.2162 \\
&& CTRL-GEN  & 0.5335 & -3.39 & 0.0107 & 0.0431 \\
&& \baby-single\ours & \underline{0.7725} & \textbf{-3.82}  & \textbf{0.0497} & \textbf{0.3152} \\
&& \baby\ours & \textbf{0.8024} & \underline{-3.68} & \underline{0.0478} & \underline{0.3056} \\

\bottomrule
\end{tabularx}
\caption{\label{cond-result} The results of conditional text generation tasks. We use \textbf{boldface} and \underline{underline} to indicate the best and the second-best performance. \baby-single indicates \baby with a \cond trained under the single condition setting, as described in Section \ref{sub:eval}. We show the natural logarithm~($\ln$) of variance, since the original scale is too small for demonstration.}
\end{center}
\end{table*}

\subsection{Baselines}
We use two semi-supervised methods, S-VAE~\cite{Kingma:14a} and CTRL-GEN~\cite{Hu:17} as our baselines. S-VAE incorporates a classifier to provide conditional distribution for unlabeled data. Note that S-VAE is originally proposed for image generation but adapted to text generation as a baseline by \citet{Hu:17}. CTRL-GEN further exploits several regularization terms to enhance the disentanglement between the structured code and the unstructured code. For a fair comparison, both the text encoder and decoder of the two baselines are the same as \glob. Furthermore, the baseline methods also exploit the same unlabeled corpus $X$ and labeled corpus $Y$ as described in the original papers.

\subsection{Models}
\baby is a model-agnostic approach, which means that both the encoders and encoders of \glob and \cond can be modified to work under different settings. Here, we describe the model architecture used in our experiments.

\paratitle{\glob.} For the encoder, we use a one-layer Bidirectional Gated Recurrent Unit (Bi-GRU) with $256$ hidden units in each direction as its encoder. Two linear Fully-Connected (FC) layers are used for re-parameteristic trick~\cite{Kingma:13}. For the decoder, we use a Transformer~\cite{Vaswani:17} ($3$ layers, $8$ heads). Additionally, we add extra positional embedding after each block, and the linearly transformed encoded vector is provided as input for each block~\cite{Brock:18}. For a fair comparison, we use the same encoder-decoder architecture for both S-VAE and CTRL-GEN.

 \paratitle{\cond.} The encoder is a two-layer FC network of 64/32 hidden units taking input in $d_g$ dimensions with an additional linear output layer of $d_c$ units. The decoder is a two-layer FC network of 32/64 hidden units taking the latent variable in $d_c$ dimensions as input with a linear output layer of $d_g$ units. 
The activation function used in the FC networks is LeakyRelu~\cite{leakyrelu}.
 
\subsection{Hyper-Parameters}
\label{hyperparam}
\paratitle{\glob.} The size of latent space $d_g$ is set to $128$. The word embedding is in $256$ dimensions and randomly initialized. 
 The output softmax matrix is tied with the embedding layer. For the adversarial classifier, we adopt two 128D hidden FC layers with LeakyRelu activation and one 1D output linear layer without bias. The balance coefficient $ \lambda $ is 20 for Yelp and 15 for News Titles. We train the WAE-GAN with Wasserstein Divergence~\cite{wu:18} to smooth the training process. The coefficient $k$ and power $p$ of Wasserstein Divergence are set to 2 and 6, respectively. During pre-training, the batch size is set to 512. Adam~\cite{Kingma:14b} with $\mathit{beta}_1=0$ is used as the optimizer. The learning rate is set to $5 \times 10^{-4}$. 
 
 \paratitle{\cond.} We set the size of latent space $d_c=20$. $\gamma$ is set to 0.1 for sentiment tasks, 0.05 for categorical tasks, and $3 \times 10^{-3}$ for length tasks. The batch size is set to 128. Adam~\cite{Kingma:14b} with $\mathit{beta}_1=0.5$ is used as the optimizer, learning rate is $3 \times 10^{-4}$ for 20K iterations. $\beta$ linearly increases from 0 to 5 in first 10K iterations.

\subsection{Evaluation Settings}
\label{sub:eval}

\paratitle{Metrics.} We evaluate the results with two metrics, accuracy and diversity. For \emph{accuracy}, we train a sentiment classifier and categorical classifier~\cite{Kim:14}, which could achieve accuracy of 90\% and 97\% on the validation set, respectively. The accuracy of \textit{length} task can be directly calculated with the word count of generated text. Plus, a model that performs well on only one condition but poorly on others is not practically useful. Thus, to measure the robustness among conditions, we calculate the variance of accuracy under all conditions in a task. For \emph{diversity}, we adopt \emph{Distinct-1} and \emph{Distinct-2}~\cite{Li:16} metrics. Distinct-1/Distinct-2 are the ratios of unique 1-gram/2-gram, respectively. A higher value indicates better diversity.
For all tasks and models, we randomly generate 10K text for each condition by greedy decoding and report the averaged results.

\paratitle{Single Condition Generation.} In a real-world scenario, the full set of conditions is not always available. When provided only a labeled set of target text (\ie $k=1$), it is not possible to learn the joint conditional space for S-VAE and CTRL-GEN any more. However, \baby can deal with that by training \textit{without} negative samples using Equation~\ref{equ-condVAE}.

\section{Experimental Results}
\label{sec:result}

\subsection{Overall Comparisons}
\label{ssec:cond-exp}

\paratitle{Accuracy.} The results of conditional text generation are listed in Table~\ref{cond-result}.
On \textit{sentiment} task, our model outperforms CTRL-GEN and S-VAE by 0.1486 and 0.129, respectively. On \textit{length} task, the accuracy of our model exceeds $95\%$, dramatically outperforming S-VAE and CTRL-GEN by 0.1124 and 0.5765 on accuracy. Notably, the performance of CTRL-GEN (0.3957) is extremely low, demonstrating the limitation of its generator-discriminator~\cite{gan:14} training process and its token-based discriminator, which is unable to discriminate text with different lengths. On \textit{topic} task, our model scores higher on accuracy than S-VAE and CTRL-GEN by 0.1094 and 0.2689, respectively. On all three tasks, \baby-single performs slightly poorer than \baby with negative samples, verifying the effectiveness of negative sampling. Furthermore, our models achieve the lowest variance on all three tasks, indicating that \baby is robust and achieves a good balance among conditions.

\paratitle{Diversity.} Diversity is a long-lasting issue lying in the field of generative models. Recent works \cite{nips17:wang,vqvae} reveal the capability of the diverse content generation with VAE-based methods. These works also conclude that VAE-based methods have better output diversity than GAN-based models. Our experimental results support this conclusion well. Particularly, CTRL-GEN suffers poor diversity, which indicates the generation of ``dull text''~\cite{Li:16}. Both S-VAE and \baby show prominently better diversity than GAN-based model, CTRL-GEN. Note that the relation between the usage of negative examples and text diversity of \baby is not statistically prominent ($p>0.05$).

\subsection{Human Evaluation}

We conduct human annotations as a complementary evaluation beyond automatic metrics. Specifically, eight individual judges are asked to rate over 200 conditional samples generated from each model and each condition. That is, for each model, a total of $4,800$ text samples are annotated. A judge needs to rate \emph{fluency} and \emph{conditionality} in the standard $1$ to $5$ scale. \emph{Fluency} measures whether the text samples are natural and fluent as real (\ie human-written) ones. \emph{Conditionality} indicates whether the generated text adheres to the given condition. 
Shown in Table~\ref{human-evaluation}, \baby achieves the best conditionality in both automatic and human evaluations on all three tasks. Meanwhile, \baby retains a satisfying fluency on \textit{sentiment} and \textit{length} tasks and obtains the best fluency on the \textit{topic} task.

\begin{table}[t!]
\small
\newcommand{\tabincell}[2]{\begin{tabular}{@{}#1@{}}#2\end{tabular}}
\begin{center}
\begin{tabular}{c|l|cc}
\toprule
Task & Method & Fluency & Conditionality \\
\midrule
\multirow{4}{*}{Sentiment} & S-VAE & 3.10 & 3.04 \\
& CTRL-GEN & \textbf{3.65} & 3.23 \\
& \baby-single & 3.54 & 3.23 \\
& \baby & 3.30 & \textbf{3.29} \\
\midrule
\multirow{4}{*}{Length} & S-VAE & \textbf{3.64} & 0.8598 \\
& CTRL-GEN & 2.53 & 0.3597 \\
& \baby-single & 3.43 & 0.9640 \\
& \baby & 3.50 & \textbf{0.9722} \\
\midrule
\multirow{4}{*}{Topic} & S-VAE & 3.31 & 2.78 \\
& CTRL-GEN & 3.09 & 2.51 \\
& \baby-single & 3.38 & 3.33 \\
& \baby & \textbf{3.45} & \textbf{3.57} \\

\bottomrule
\end{tabular}
\caption{\label{human-evaluation} Human evaluation results. Note that since the \textit{length} task is objectively defined, we copy the accuracy results from Table \ref{cond-result}.}
\end{center}
\end{table}

\subsection{Training Costs}
\label{traincosts}
To measure the efficiency of proposed methods, we report the training time and the number of parameters of S-VAE, CTRL-GEN and \baby in Table~\ref{effi}. We train the models on a single Nvidia GTX 1080 GPU and report the training time until the convergence of each model. \glob has the same size of S-VAE but only needs to be trained once and does not require a full retraining when a new condition added. Also, \cond, which learns to transform between the global latent space and the conditional latent space, only has 22K parameters and can be trained within about one minute.

\begin{table}[t]
\small
\newcommand{\tabincell}[2]{\begin{tabular}{@{}#1@{}}#2\end{tabular}}
\begin{center}
\begin{tabular}{l|cc}
\toprule Method & \#~Training Params & Training Time \\
\midrule
S-VAE & 6.5M & 1.4h \\
CTRL-GEN & 8.5M & 3.5h \\
\midrule
\glob & 6.5M & 1.2h (only once) \\
\cond & 22K & 64s \\
\bottomrule
\end{tabular}
\caption{\label{effi} Average numbers of parameters and time costs for training.}
\end{center}
\end{table}

\begin{table}[t]
\small
\begin{center}
\begin{tabular}{c|l|cc}
\toprule
Task & Method & Acc. & Distinct-1/2 \\
\midrule
\multirow{3}{*}{Sentiment} & Fine-tuning & 0.5319 & 0.0281~/~\textbf{0.2845} \\
& \baby-single & 0.7832 & 0.0350~/~0.2568 \\
& \baby & \textbf{0.8484} & \textbf{0.0356}~/~0.2627\\
\midrule
\multirow{3}{*}{Length} & Fine-tuning & 0.9456 & 0.0340~/~\textbf{0.2923} \\
& \baby-single & 0.9640 & \textbf{0.0375}~/~0.2549 \\
& \baby & \textbf{0.9722} & 0.0372~/~0.2538\\
\bottomrule
\end{tabular}
\caption{\label{vsfinetune} The comparisons of fine-tuned \glob with the full \baby on the two tasks of Yelp dataset.}
\end{center}
\end{table}

\begin{table}[t!]
\small
\newcommand{\tabincell}[2]{\begin{tabular}{@{}#1@{}}#2\end{tabular}}
\begin{center}
\begin{tabular}{c|ccc}
\toprule
$\beta$ &  Accuracy &  Distinct-1 & Distinct-2 \\ 
\midrule
0.0 & 1.0000  & 0.0001 & 0.0001  \\
2.0 & \textbf{0.9938} & 0.0256 & 0.1629 \\
5.0 & 0.9908  & 0.0301 & 0.2112\\
10.0 & 0.9875  & \textbf{0.0324} & \textbf{0.2370} \\
\bottomrule
\end{tabular}
\caption{\label{diff-c} The impact of different $\beta$ on long text generation task. }
\end{center}
\end{table}

\begin{table*}[t]
\small
\begin{center}
\begin{tabular}{l|l|l}
\toprule
\textbf{Task} & \textbf{Condition} & \textbf{Generated Examples}\\
\midrule
\multirow{2}{*}{Sentiment} & Positive & The services are friendly, fast.\\
& Negative & The egg drop soup was old and tasted like feet. \\
\midrule
\multirow{3}{*}{Length} & Short & Great pricing! \\
& Medium & I refused to work with you and this place. \\
& Long & And this made me feel uncomfortable and the prices aren't right.\\
\midrule

\multirow{3}{*}{Topic} & Business & FDA Approves New Case of E-cigarettes\\
& Health & Ebola : Virus Spreads in the US\\
& Entertainment & Surprise Birthday: The Guys of the Cast of Disney Parks \\
\bottomrule
\end{tabular}
\caption{\label{case-study} Some conditional examples generated by \baby for qualitative analysis (cherry-picked).}
\end{center}
\end{table*}

\begin{table}[t]
\small
\newcommand{\tabincell}[2]{\begin{tabular}{@{}#1@{}}#2\end{tabular}}
\begin{center}
\resizebox{\columnwidth}{!}{
\begin{tabular}{l}
\toprule
\textbf{Generated Examples}\\
\midrule
\textbf{S-VAE }\\
Chinese State Media: 17 Miners Trapped Underground \\
Huge Increases in Obamacare Premiums Are Coming\\ Herbalife Ltd. (HLF) Probe Earns Bill Ackman Back Millions \\
\midrule
\textbf{CTRL-GEN}\\
Pfizer's Astrazeneca's Astrazeneca Bid for Astrazeneca \\
FDA's New Drug to Treat Migraines \\
Pfizer to Acquire Seragon in \$42.9B \\

\midrule

\textbf{\baby} \\
Despite Highway Crisis, Many Worries Remain on US Oil Exports \\
Lululemon: Digital Sales Surge in 1Q Net Income, Revenue \\
Crisis of Market: US Stocks Climb; Nike Jumps \\
\bottomrule
\end{tabular}
}
\caption{\label{diverse-study} Some generated conditional examples under condition \textit{Business} (randomly sampled).}
\end{center}
\end{table}

\begin{table}[t]
\small
\newcommand{\tabincell}[2]{\begin{tabular}{@{}#1@{}}#2\end{tabular}}
\begin{center}
\resizebox{0.9\columnwidth}{!}{
\begin{tabular}{l}
\toprule
\textbf{Failed Examples}\\
\midrule
\textbf{Grammatical} \\
Eat the service! \\
In addition, this location sucks it is. \\
Star Wars 7 will include US production on set \\
\midrule
\textbf{Conditional} \\
\textit{(Negative)} I was shocked that this is what I needed.  \\
\textit{(Long)} Are you actually drunk outside? \\
\textit{(Business)} Michael Jackson's New Album `Xscape' \\

\bottomrule
\end{tabular}
}
\caption{\label{error-analysis} Some failed examples (cherry-picked).}
\end{center}
\end{table}

\subsection{\cond vs. Fine-Tuning}
As a natural baseline, the conditional generation can also be done by directly fine-tuning \glob on each condition. Shown in Table \ref{vsfinetune}, despite the fact that it is not computationally efficient and saving the full weights is undesirable for industrial applications when the model is large (\eg GPT-2~\cite{gpt2}), both \cond trained with and without negative samples significantly outperform a directly fine-tuned \glob on accuracy.

\subsection{Effect of Hyper-parameter $\beta$}
\label{ssec:para-c}

Since $\beta$ is an important hyper-parameter for \baby, we test $\beta \in \{0, 2, 5, 10 \}$ on the long text generation task. From the results in Table~\ref{diff-c}, we find that $\beta$ controls the balance between diversity and accuracy. Specifically, when $\beta$ is too large, more diverse samples could be generated, but the accuracy may be sacrificed slightly. On the contrary, when $\beta$ is too small, the accuracy could climb to a higher value, but meanwhile, the diversity drops drastically. Empirically, we find that $\beta=5$ is an appropriate value for all tasks.

\section{Case Study}

We select some generated conditional text of each condition in Table~\ref{case-study}. As shown in the table, our proposed \baby is capable of generating realistic conditional text. Also, shown in Table \ref{diverse-study}, on \textit{topic} task, we randomly select some examples from the output of each model. The output of S-VAE seems to be diverse but is poorly conditioned. CTRL-GEN suffers an obvious diversity issue, which makes it repeatedly output similar text.

For the error analysis, we pick some failed examples of \baby in Table~\ref{error-analysis}. We categorize the errors into two main classes. (1) \textbf{Grammatical}. Grammatical problems are common in NLG. As we analyze, this kind of errors can be mitigated with a deeper encoder and decoder with even more unlabeled data for pre-training. (2) \textbf{Conditional}. Conditional errors are of great interest to us since they lie in our focus. We choose three typical errors and list them in Table \ref{error-analysis}. In the first sentence, ``shocked'' is a subtle word which may indicate either positive or negative sentiment depending on the context. Thus, with a greedy decoding strategy, it may be incorrectly decoded into the other polarity. We believe this kind of errors could be fixed with more elaborate decoding strategies (\eg Weighted Decoding~\cite{see2019makes}). In the second sentence, the length is limited by the nature of an interrogative sentence. As a linguistic fact, an interrogative sentence often has fewer words than a declarative sentence. In the third sentence, we remark an overlapping problem between classes. Some topics (\eg music album) may appear in both business and entertainment news. In some way, these samples can also be considered as correctly conditioned ones, which highlights the importance of a fine-grained human evaluation on this task.

\section{Conclusion}

In this paper, we present a novel \baby framework for flexible conditional text generation, which decouples the text generation module from the condition representation module. The extensive experiments demonstrate the superiority of the proposed \baby against the existing alternatives on conditionality and diversity while allowing new conditions to be added without a full retraining.

\section*{Acknowledgments}
We are grateful for the insightful comments from the anonymous reviewers.
We would like to especially thank Daya Guo for his help and suggestions. This research was supported by National Natural Science Foundation
of China (No. 61872278). Chenliang Li is the corresponding author.

\bibliography{ppvae}
\bibliographystyle{acl_natbib}

\end{document}